\documentclass{article}
\usepackage{spconf,amsmath,graphicx,hyperref}
\usepackage{subfigure}
\usepackage{dsfont}
\usepackage{bbm}
\usepackage{multirow} 
\usepackage{booktabs}

\title{PERSONALIZED CELL SEGMENTATION: BENCHMARK AND FRAMEWORK FOR REFERENCE-GUIDED CELL TYPE SEGMENTATION}
\name{Bisheng Wang$^{1}$, 
Jaime S. Cardoso$^{2}$,  
and Lin Wu$^{3}$
}

\address{$^{1}$ INESC TEC, Portugal
$^{2}$ INESC TEC, Faculdade de Engenharia, Universidade do Porto, Portugal \\
$^{3}$ James Watt School of Engineering, University of Glasgow, United Kingdom
}
\begin{document}
%
\maketitle
\begin{abstract}
Accurate cell segmentation is critical for biological and medical imaging studies. Although recent deep learning models have advanced this task, most methods are limited to generic cell segmentation, lacking the ability to differentiate specific cell types.
In this work, we introduce the Personalized Cell Segmentation (PerCS) task, which aims to segment all cells of a specific type given a reference cell. To support this task, we establish a benchmark by reorganizing publicly available datasets, yielding 1,372 images and over 110,000 annotated cells.
As a pioneering solution, we propose PerCS-DINO, a framework built on the DINOv2 backbone. By integrating image features and reference embeddings via a cross-attention transformer and contrastive learning, PerCS-DINO effectively segments cells matching the reference.
Extensive experiments demonstrate the effectiveness of the proposed PerCS-DINO and highlight the challenges of this new task. We expect PerCS to serve as a useful testbed for advancing research in cell-based applications.
\end{abstract}
\section{Introduction}
\label{sec:intro}

Cell segmentation~\cite{petukhov2022cell, ma2024multimodality}, the task of delineating individual cells or their substructures (e.g., nuclei, cytoplasm, membranes) from microscopic images, is a fundamental step in biological image analysis. It underpins downstream tasks such as quantitative analysis and decision-making, and critically determines the reliability and reproducibility of biological discoveries. Therefore, accurate cell segmentation is indispensable across biomedical research, drug discovery, and high-throughput biological data processing~\cite{driscoll2021data}.

Recent advances in deep learning and the emergence of foundation models have driven remarkable progress in general-purpose cell segmentation. Approaches such as Cellpose~\cite{stringer2021cellpose}, CellSAM~\cite{pachitariu2025cellpose}, and Cellpose-SAM~\cite{israel2025cellsam} leverage large pre-trained models to achieve impressive performance across diverse datasets. However, these methods are trained for generic segmentation, lacking the flexibility to handle unseen or user-specified cell types without retraining or extensive manual annotation. 
For example, in many applications it is important to segment specific cell types (e.g., tumor cells or rare stem cells), which current frameworks cannot efficiently address.
Personalized segmentation~\cite{zhang2023personalize, siddiqui2025towards} offers an alternative by segmenting all targets belonging to the same class as a given reference example. However, existing frameworks~\cite{zhang2023personalize, liu2023matcher} are designed for natural images, where objects are relatively sparse, distinct, and large. In contrast, cell images involve small sizes, dense adhesion, morphological variability, and heterogeneous modalities (e.g., fluorescence, bright-field), making direct transfer particularly difficult.

To address these limitations, we introduce Personalized Cell Segmentation (PerCS), which aims to segment all cells of a target type within an image, given only a single annotated reference cell of any type. To facilitate research in this direction, we reorganize multiple public cell segmentation datasets~\cite{stringer2021cellpose, israel2025cellsam, dietler2020convolutional, spahn2022deepbacs, kim2014yeastnet} and construct a new benchmark containing 1372 images and over 110,000 annotated cells spanning 108 cell types. The dataset covers cells from multiple imaging sources, varying densities, and diverse modalities (e.g., fluorescence, bright field, histology), capturing diverse and realistic challenges in bioimage analysis.

Building on this task, we propose a strong baseline framework, PerCS-DINO, which leverages DINOv2~\cite{oquab2023dinov2} as the backbone for discriminative feature extraction. To enable personalized segmentation, we design a reference-guided cross-attention transformer~\cite{vaswani2017attention}, allowing the reference cell features to interact with image features and guiding the model to automatically learn feature similarity for personalized cell segmentation. In addition, a cell-level supervised contrastive loss~\cite{khosla2020supervised} is integrated to enhance inter-class separation among different types of cells. Experiments on our PerCS benchmark and PerPanNuke datasets demonstrate the strong performance and generalization capability of PerCS-DINO.


\section{Task Setting and Datasets}
\label{sec:task}

\begin{figure}
\centering
\includegraphics[width=0.85\linewidth]{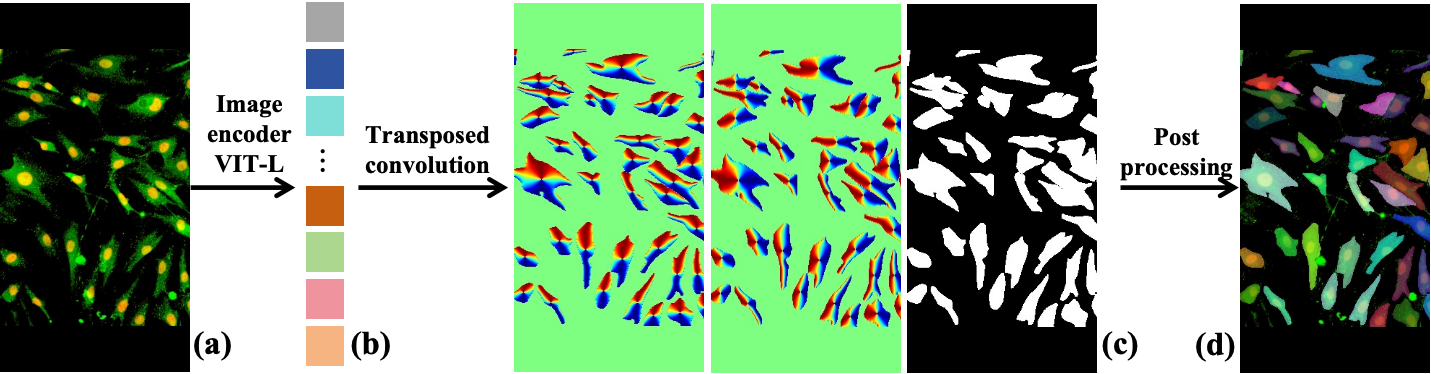}
\caption{Framework of generic method, Cellpose-SAM~\cite{pachitariu2025cellpose}. (a) Input images. (b) Image features. (c) Horizontal and vertical vector flows and binary cell mask. (d) Cell mask. }
\label{fig:generic_cellseg}
\end{figure}

Fig.~\ref{fig:generic_cellseg} illustrates the traditional cell segmentation framework of Cellpose-SAM~\cite{pachitariu2025cellpose}. An input image is first encoded into patch-wise feature embeddings, which are then upsampled through a transposed convolution layer to generate pixel-wise vector flows. These vector flows are finally transformed into cell masks via post-processing.
Although Cellpose-SAM demonstrates strong generic segmentation performance, it lacks the ability to distinguish between different cell categories. In practice, microscopy images often contain multiple cell types, and isolating a particular category of interest is often required. Achieving this with existing frameworks requires retraining, which is labor intensive and inflexible.

\subsection{Personalized Cell Segmentation (PerCS)}
\label{ssec:PerCS}

To address this limitation, we introduce the task of Personalized Cell Segmentation (PerCS). As illustrated in Fig.~\ref{fig:framework}, given a query image and a reference cell (which may be sampled either from the query or from another image), the objective of PerCS is to segment all cells in the query image that belong to the same category as the reference. Unlike generic cell segmentation, PerCS supports user-driven and category-specific segmentation without retraining.

Several studies have investigated personalized segmentation~\cite{zhang2023personalize, siddiqui2025towards} in the context of natural images. However, prior methods are typically for semantic segmentation or sparse, large-scale instances. These approaches do not transfer well to cell images, where objects are small, densely clustered, and morphologically diverse. This motivates the development of a dedicated framework for PerCS in bioimage analysis.

\subsection{Datasets}
\label{ssec:Datasets}

\subsubsection{PerCS Dataset}

\begin{figure}
\centering
\includegraphics[width=0.8\linewidth]{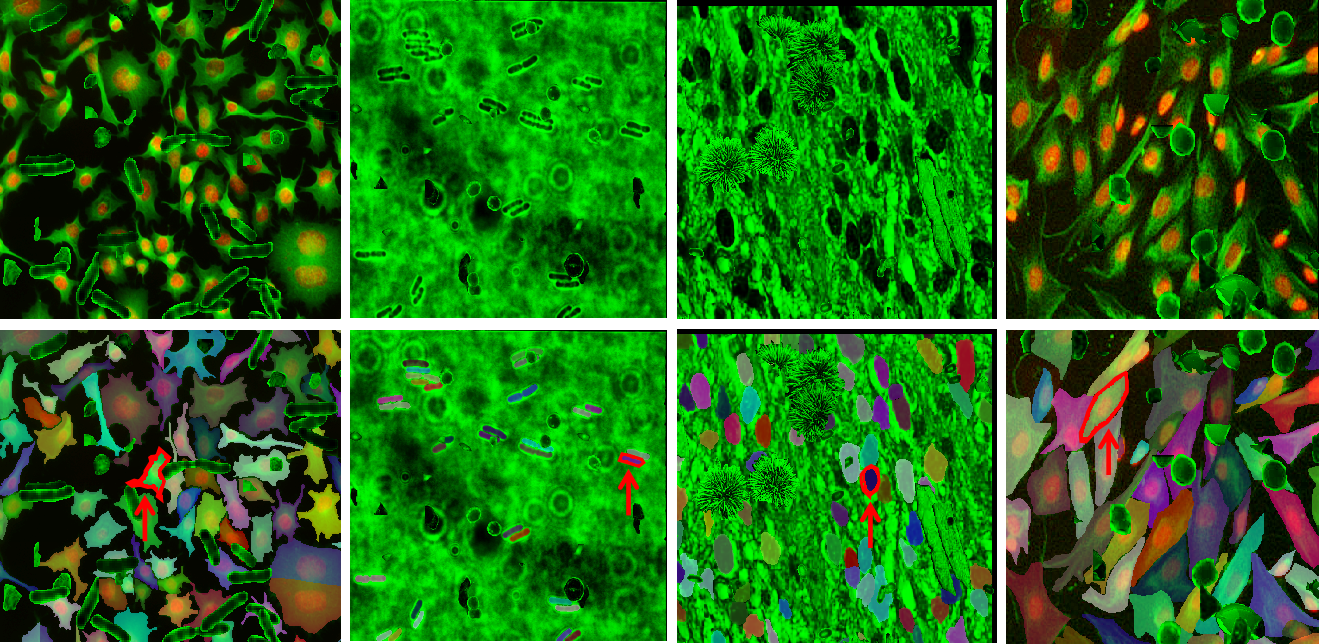}
\caption{Sample images (top row) and corresponding colored annotation masks (bottom row) from the PerCS dataset. Reference targets are outlined in red in the annotation masks.}
\label{fig:examples}
\end{figure}

To facilitate research on PerCS, we reorganize multiple public cell segmentation datasets (including Cellpose~\cite{stringer2021cellpose}, YeaZ~\cite{dietler2020convolutional}, YeastNet~\cite{kim2014yeastnet}, BriFiSeg~\cite{mathieu2022brifiseg}, CellSAM~\cite{israel2025cellsam}) and construct a comprehensive benchmark consisting of 1372 images and over 110,000 annotated cells. The dataset covers cells of different shapes, varying densities, and diverse modalities (e.g., fluorescence, bright-field, histology), capturing diverse and realistic challenges in bioimage analysis. 
Unlike existing resources that typically provide images without explicit category-level organization, we perform a rigorous manual reorganization to reclassify all images into 108 distinct types. This classification is based on distinct morphological features and available biological metadata provided by the original sources. To ensure label reliability, the categorization process underwent independent verification, filtering out ambiguous samples. This fine-grained categorization enables targeted evaluation of cross-class generalization. Some example images are shown in Fig.~\ref{fig:examples}.

The manually classified dataset is then divided into three subsets: train, test, and novel set. The training set (1,186 images) and test set (105 images) share overlapping cell types, while the novel set (81 images) contains unseen types, serving as a benchmark for evaluating generalization to previously unobserved categories. To better approximate real microscopy scenarios where multiple cell types co-occur, we additionally generate mixed-type images: for each image, cells from different types are randomly pasted from other images. This augmentation is applied dynamically during training and fixed during testing to ensure consistency and fairness.

\subsubsection{PerPanNuke Dataset}
In addition to our main PerCS benchmark, we further construct a personalized PanNuke dataset (PerPanNuke) based on the PanNuke dataset~\cite{gamper2019pannuke}. Compared to our main PerCS benchmark, PerPanNuke introduces additional challenges due to the coexistence of multiple nuclear types within the same histology image, which often exhibit highly similar morphological characteristics. This makes distinguishing between categories substantially more difficult.

PanNuke provides high-quality multi-class annotations for five cell nucleus types. Instead of treating it as a conventional multi-class segmentation task, we reformulate it under the PerCS setting by strictly mapping the original class labels to the reference-guided protocol: given a reference nucleus of a specified category, the model is required to segment all nuclei of the same type within the image. We follow the official split protocol of PanNuke, using folds 1 and 2 for training and fold 3 for evaluation.

For reference selection in both datasets, we adopt consistent strategies to ensure fairness and reproducibility. During training, in each iteration, a reference cell (or nucleus) is randomly sampled from the training set. During testing, a fixed reference target is predefined for every image and kept identical across all methods, thereby enabling fair comparison under the same evaluation protocol.

\begin{figure}   
\centering
\includegraphics[width=0.7\linewidth]{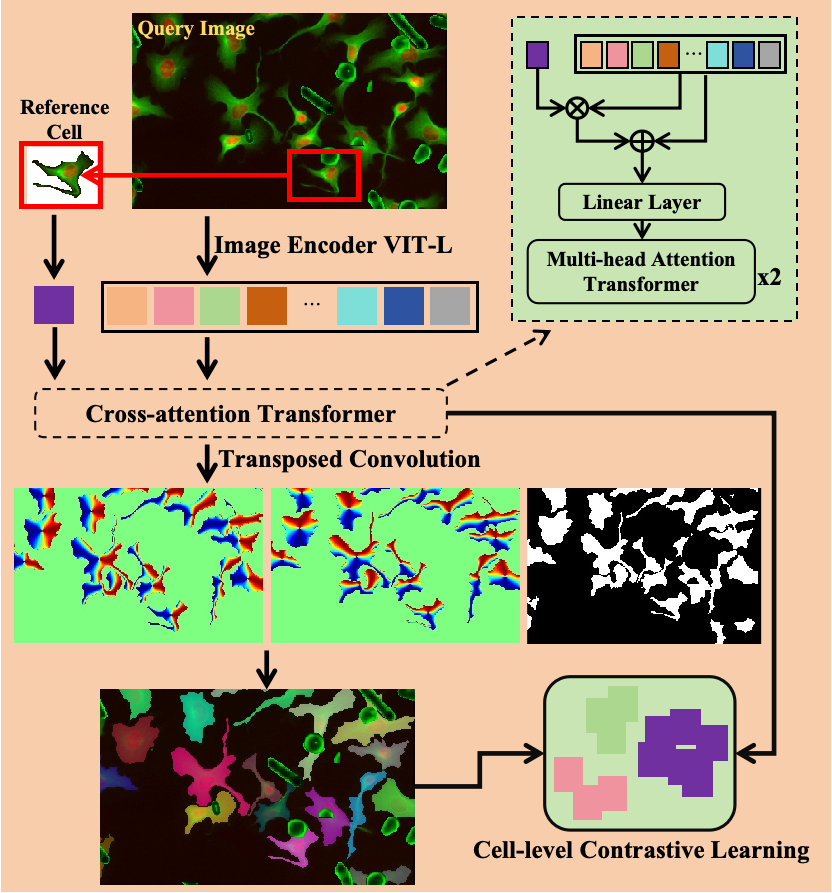}
\caption{Framework of our proposed method, PerCS-DINO.}
\label{fig:framework}
\end{figure}

\section{Baseline Framework}
\label{sec:method}

\subsection{Adaptation of Existing Works}
\label{ssec:existing work}
Since no prior methods directly address our task, we adapt two representative approaches as baselines.

\textbf{Personalized SAM.} Since the original personalized SAM~\cite{zhang2023personalize} supports only single-object segmentation, we extend it by selecting 100 candidate points that are both spatially separated and visually similar to the reference target. These points are then taken as prompts to segment cells. Non maximum suppression (NMS) is adopted to remove duplicated segmentation.

\textbf{Cellpose-SAM.} Cellpose-SAM~\cite{pachitariu2025cellpose} is a method of generic cell segmentation and lacks the ability to differentiate cell categories. We adopt the similarity-based filtering strategy from personalized SAM~\cite{zhang2023personalize} to filter its predictions which have different cell types from referent cell. Note that since the novel split partially overlaps with the Cellpose-SAM training data, we do not report the results of Cellpose-SAM on this split. 

\subsection{PerCS-DINO}
\label{ssec:PerCS-DINO}

As illustrated in Fig.~\ref{fig:framework}, we propose PerCS-DINO, a personalized cell segmentation framework tailored for dense and heterogeneous cellular images. 
The framework consists of three main components: a DINOv2-based backbone for feature extraction, a reference-guided cross-attention transformer for feature interaction, and a cell-level contrastive learning module for inter-class discrimination. 

\textbf{Backbone with DINOv2.} We adopt DINOv2 ViT-L as the backbone to extract global image representations. Unlike the ViT encoder in SAM~\cite{kirillov2023segment}, DINOv2 yields more discriminative features, which is critical to distinguish subtle differences between cell types.

\textbf{Reference-guided cross-attention and decoding.}  
Given the feature map of the query image $f_{q} \in \mathds{R}^{B \times HW \times D}$ and the embedding of a reference cell $f_{ref} \in \mathds{R}^{B \times 1 \times D}$ (where $B$, $HW$, and $D$ denote batch size, the flattened spatial resolution, and the feature dimension, respectively), we first compute a cosine similarity map $f_{s} \in \mathds{R}^{B \times HW \times 1}$. The concatenated features $[f_{q}; f_{s}]$ are projected through a linear layer to 256 dimensions and then processed by a two-layer multi-head attention transformer with four heads. This module enables an effective interaction between reference and query features, allowing the network to propagate category-specific cues across the image.  
The fused representation is subsequently decoded through transposed convolutions into a three-channel output. Two channels encode vector flow fields, which represent the spatial gradients directing each pixel toward the topological center of its corresponding cell instance~\cite{stringer2021cellpose}, while the third channel produces a probability map indicating the likelihood that each pixel belongs to the same category as the reference cell. The computation of ground truth vector flows and the subsequent post-processing for mask reconstruction follow the Cellpose framework~\cite{stringer2021cellpose}.

\textbf{Cell-level supervised contrastive learning.}  
While reference-guided cross-attention improves patch-level similarity modeling, it is insufficient to ensure robust category discrimination. To address this, we incorporate a supervised contrastive learning objective~\cite{khosla2020supervised} at the cell level to encourage embeddings of the same category to cluster together while pushing apart embeddings of different categories. After cross-attention layer, pixel embeddings belonging to the same cell are aggregated via global pooling to form cell-level representations.  
Formally, let $\mathbf{z}_i$ denote the embedding of cell $i$, normalized to unit length, and let $y_i$ be its class label. The supervised contrastive loss for a batch of $N$ cells is defined as:
\begin{equation}
\mathcal{L}_{\text{SCL}} = \sum_{i=1}^{N} \frac{-1}{|P(i)|} \sum_{p \in P(i)} \log \frac{\exp(\mathbf{z}_i \cdot \mathbf{z}_p / \tau)}{\sum_{a=1}^{N} \mathbbm{1}_{[a \neq i]} \exp(\mathbf{z}_i \cdot \mathbf{z}_a / \tau)},
\end{equation}
where $P(i) = \{p \in \{1,\ldots,N\} \setminus \{i\} \, | \, y_p = y_i\}$ is the set of positive samples sharing the same class as $i$, $\tau$ is a temperature hyper-parameter, and $\cdot$ denotes dot-product similarity.  

Finally, the overall training loss combines three objectives: (i) a mean squared error (MSE) loss $\mathcal{L}_{\text{MSE}}$ on vector flow prediction, (ii) a binary cross-entropy loss with logits $\mathcal{L}_{\text{BCE}}$ for pixel-wise category classification, and (iii) the supervised contrastive loss $\mathcal{L}_{\text{SCL}}$. The full loss function is:
\begin{equation}
\mathcal{L} = \mathcal{L}_{\text{MSE}} + \mathcal{L}_{\text{BCE}} + \mathcal{L}_{\text{SCL}}.
\end{equation}

\begin{table*}[t]
\small
\centering
\caption{Comparison of PerCS-DINO with baseline methods under different IoU thresholds.}
\begin{tabular}{l c | ccc | ccc | ccc}
\toprule
\multirow{2}{*}{Methods} & \multirow{2}{*}{IoU} 
& \multicolumn{3}{c|}{PerCS Test} & \multicolumn{3}{c|}{PerCS Novel} & \multicolumn{3}{c}{PerPanNuke} \\
& & AP & P & R & AP & P & R & AP & P & R \\
\midrule
PerSAM       & 0.3 & 0.241 & 0.513 & 0.313 & 0.249 & 0.540 & 0.317 & 0.151 & 0.186 & 0.445 \\
Cellpose-SAM  & 0.3 & 0.465 & 0.618 & 0.654 & - & - & - & 0.368 & 0.433 & 0.708  \\
\textbf{PerCS-DINO}   & 0.3 & \textbf{0.739} & \textbf{0.837} & \textbf{0.862} & \textbf{0.660} & \textbf{0.758} & \textbf{0.836} & \textbf{0.395} & \textbf{0.442} & \textbf{0.785} \\
\midrule
PerSAM       & 0.5 & 0.153 & 0.350 & 0.214 & 0.171 & 0.394 & 0.231 & 0.112 & 0.142 & 0.340 \\
Cellpose-SAM  & 0.5 & 0.428 & 0.583 & 0.617 & - & - & - & 0.334 & 0.403 & 0.659 \\
\textbf{PerCS-DINO}   & \textbf{0.5} & \textbf{0.655} & \textbf{0.780} & \textbf{0.804} & \textbf{0.560} & \textbf{0.685} & \textbf{0.755} & \textbf{0.364} & \textbf{0.418} & \textbf{0.740} \\
\bottomrule
\end{tabular}
\label{tab:results}
\end{table*}

\section{Experiments}
\label{sec:experiments}

\subsection{Implementation Details}

We adopt DINOv2-Large~\cite{oquab2023dinov2} as the backbone. During training, all images are randomly cropped to a fixed resolution of $336 \times 336$. This resolution is explicitly chosen to satisfy the patch-size alignment constraints of DINOv2 (multiples of 14) while optimizing the trade-off between receptive field coverage and computational efficiency. Data augmentation includes random horizontal and vertical flips. During inference, we employ a sliding-window strategy for larger images: patches of size $336 \times 336$ are extracted with a stride of 168, vector flows are predicted for each patch, and overlapping regions are averaged to obtain the final segmentation. For the PerPanNuke dataset, the input resolution is kept at $224 \times 224$, consistent with the original image size.

The model is trained for 60 epochs with a batch size of 1 using the Adam optimizer. The initial learning rate is $1 \times 10^{-5}$ with a weight decay of 0.1, and is reduced by a factor of 10 during the last 10 epochs. The temperature $\tau$ in the contrastive loss is fixed to 0.1, empirically set to balance hard example mining with feature space smoothing. All experiments are conducted on a single NVIDIA A100 GPU.

\subsection{Evaluation Metrics}

Segmentation performance is evaluated with instance-level matching, where predicted masks are paired with ground-truth cells via the Hungarian algorithm. A match is considered a true positive (TP) if the intersection-over-union (IoU) exceeds a threshold; unmatched predictions are false positives (FP), and unmatched ground-truth cells are false negatives (FN).

We report $\text{AP}$, $\text{Precision (P)}$, and $\text{Recall (R)}$ under IoU thresholds of 0.3 and 0.5 to capture both tolerant and stricter criteria: $\text{AP} = \frac{TP}{TP + FP + FN}$, $\text{P} = \frac{TP}{TP + FP}$, $\text{R} = \frac{TP}{TP + FN}$.


\subsection{Results Comparison}

Table~\ref{tab:results} reports the performance of three methods in the PerCS test, Novel split, and the PerPanNuke dataset.  

PerSAM performs poorly across all datasets, indicating that direct transfer from natural images to dense biological scenarios is ineffective. In contrast, Cellpose-SAM achieves higher recall, benefiting from its pretraining on large-scale cell images. However, as it lacks the ability to differentiate cell categories, it suffers from substantial false positives, even after applying cell-level similarity filtering.  

Our PerCS-DINO, trained specifically for the personalized cell segmentation task, surpasses both baselines and achieves the best performance on the PerCS test set, reaching AP of 73.9\% and 65.5\% at IoU threshold of 0.3 and 0.5, respectively. On the PerCS novel split, PerCS-DINO also demonstrates competitive segmentation quality, providing strong evidence of its zero-shot transferability to unseen cell types. 
PerCS-DINO also attains the highest AP on PerPanNuke, although performance remains limited by the high morphological similarity among nucleus categories in histopathology images, underscoring the intrinsic difficulty of the PerCS task and motivating future improvements in reference modeling.


\subsection{Difficulty Analysis}

Despite the encouraging results of our method, two key challenges remain. First, segmentation masks lack refinement: downsampling in the backbone reduces spatial details, leading to a clear performance drop under stricter IoU thresholds (e.g., 0.5 vs. 0.3). Second, the framework relies on a single reference cell, which is insufficient given the large morphological variability across cell types. This limitation is particularly evident in PerPanNuke, where visually similar nuclei often cause false positives despite high recall. Future work should therefore focus on more discriminative feature learning or the integration of multiple positive and negative references to enhance robustness.

\section{Conclusion}
\label{sec:conclusion}

We introduced the task of Personalized Cell Segmentation (PerCS), which aims to segment all instances of a target cell type given only a single reference cell. To support this task, we constructed a benchmark by reorganizing multiple public datasets, covering over 110k annotated cells across 108 categories and diverse imaging modalities. Finally, we proposed PerCS-DINO, a baseline framework that combines DINOv2 features, reference-guided cross-attention, and cell-level contrastive learning. Experiments on PerCS and PerPanNuke demonstrate its effectiveness while highlighting the challenges of personalized bioimage analysis.

\textbf{Acknowledgments.}
This work is co-financed by Component 5 - Capitalization and Business Innovation, integrated in the Resilience Dimension of the Recovery and Resilience Plan within the scope of the Recovery and Resilience Mechanism (MRR) of the European Union (EU), framed in the Next Generation EU, for the period 2021 - 2026, within project NewSpacePortugal, with reference 11

{\small
\bibliographystyle{IEEEbib}
\bibliography{refs}
}

\end{document}